\documentclass[sigconf,nonacm]{acmart}
\AtBeginDocument{%
  }

\begin{document}
\title{RxnCLF: Contrastive Transformation-Aware Reaction Foundation Model for Improved Reactivity Prediction}

\author{Yiting Zheng}
\orcid{1234-5678-9012}
\affiliation{%
  \institution{Discovery Chemistry, Merck \& Co., Inc.}
  \city{Boston}
  \state{MA}
  \country{USA}
}

\author{Cheng Fang}
\correspondingauthor
\affiliation{%
  \institution{Discovery Chemistry, Merck \& Co., Inc.}
  \city{Cambridge, MA}
  \country{USA}}

\author{Anthony Donofrio}
\orcid{1234-5678-9012}
\affiliation{%
  \institution{Discovery Chemistry, Merck \& Co., Inc.}
  \city{Boston}
  \state{MA}
  \country{USA}
}

\author{Haote Li}
\affiliation{%
  \institution{Discovery Chemistry, Merck \& Co., Inc.}
  \city{Boston}
  \state{MA}
  \country{USA}
}

\begin{abstract}

Reaction yield prediction remains challenging because labeled data are scarce and reaction space is both combinatorially large and sparsely populated, limiting the generalization of existing reaction representations. String-, fingerprint-, and graph-based reaction encodings only partially capture chemical transformations, making accurate prediction difficult for reactions with complex substrates.

We propose reaction contrastive learning foundation (RxnCLF), a self-supervised contrastive framework for reaction representation learning. RxnCLF is built on a condensed reaction graph (CRG) that unifies reactant and product information into a single graph, enabling the model to learn explicit and enriched transformation structure rather than disconnected graphs. Pretrained on 1.7 million Pistachio reactions, RxnCLF learns a compact and continuous latent space that captures both reaction-center features and broader side chain contexts, making it  transformation-aware and chemically interpretable. Fine-tuned on multiple yield prediction benchmarks, including Buchwald-Hartwig, Pd-catalyzed BH coupling, and proprietary HTE C-N coupling and amide formation datasets, RxnCLF consistently outperforms graph and sequence-based baselines, improving $R^2$ and achieving the best performance overall.

Our results highlight the promise of CRG-based RxnCLF as a scalable reaction foundation model, with the potential to generalize across broader reaction spaces and support diverse downstream reaction informatics tasks, including regioselectivity prediction, enantioselectivity prediction, and reaction condition optimization.
\end{abstract}


\maketitle

\section{Introduction}
Predicting reaction properties such as yield, reaction enthalpy, activation energy, and kinetic rate is a core problem in chemistry with direct impact on synthesis planning, reaction optimization, and drug discovery~\cite{when_ml_meets_mol, predictive_chemistry, delta2_machine_learning}. Despite substantial progress in machine learning methods for molecular property predictions, applying them to reaction remains challenging as it depends not only on the structures of participating molecules but also on the specific transformation that occurs between them. Thus, designing representations that faithfully encode reaction semantics becomes critical.

Most existing methods represent reactions using SMILES strings~\cite{drfp, robust_ood_bh, exploring_bert, yieldbert}, reaction fingerprints~\cite{drfp, reaction_fp}, or graphs~\cite{graphrxn, graph_transformation_policy_network, enzyme_prediction_hypergraph}. While these approaches are effective at capturing structure for reaction participants, they rarely model the transformation-level information that governs reaction behavior~\cite{bioinformatics}. Additionally, inferring reaction-center and bond-change patterns based on molecular input may compromise their ability to capture broader reaction information. 

BERT-based encoders, exemplified by YieldBERT, have become widely used for reaction yield predictions~\cite{yieldbert, robust_ood_bh, exploring_bert}. Built on RxnFP, YieldBERT captures the latent structure of reaction space, demonstrating the potential of representation learning. However, they remain limited by tokenization strategy, computational cost, and the fact that linearized molecular formats such as SMILES transform molecular graphs into sequences, introducing a sequential bias that may hinder the modeling of complex topological dependencies ~\cite{retrosynthesis_tu, gotta_be_safe}. In contrast, graph neural networks offer promising approach by directly modeling atom- and bond-level information ~\cite{gnn_mol, chemprop}. Although reactions represented with disconnected graphs from individual participants are explored in the past ~\cite{graphrxn, syntemp}, generalizing them to represent reactions is not obvious. This is because a reaction is a transformation over multiple reactants and products rather than a single molecular graph~\cite{doi:10.1021/acs.jcim.1c00975}. 

Another challenge for reaction property prediction is that reaction space is combinatorially large, while experimentally labeled data is scarce, costly, and concentrated in only a small portion of accessed space ~\cite{ood_bert, foundation_model_transfer, learning_from_models_beyond_finetuning}. In this regime, supervised models trained on existing benchmarks cannot extrapolate towards regions of scarce reaction chemistries ~\cite{molclr}. Foundation models offer a principled solution by leveraging large-scale unlabeled or weakly labeled reaction data to learn transformation patterns that transfer across tasks and reaction conditions. 

To address these challenges, we propose RxnCLF, a contrastive learning (CL) framework built on a reaction foundation model that constructs a chemically meaningful and transformation-aware latent space from the condensed reaction graph (CRG). CRG unifies reactant and product information in a single graph, explicitly encoding bond formation/cleavage, and conserved substructures. It catalyzes a more faithful and scalable representation of reaction transformations than disconnected graph or string-based encodings~\cite{doi:10.1021/acs.jcim.1c00975, sr_smiles, varnek2005}. Pretrained with CL on large-scale reaction data, RxnCLF pulls similar reaction graphs (CRG) closer, enabling transferable representations that support inter-dataset generalization with sparse labels.

Extensive experiments show that features derived from RxnCLF consistently improves yield predictions, especially under low-data and out-of-distribution settings. These results further demonstrate that learning a transformation-aware reaction space is both chemically meaningful and practically effective. This paper’s contributions are summarized as follows:

\begin{itemize} \item Introducing a reaction foundation model, which constructs chemically interpretable reaction latent space to capture reaction-center features and side-chain context in a unified tensor.
\end{itemize}
\begin{itemize} \item Demonstrating that this representation contributes to consistent improvements in downstream yield prediction across both academic and industrial benchmarks, highlighting the practical value of interpretable reaction representations with sparse labels.
\end{itemize}

\section{Methods}

\subsection{CRG generation} For reaction representation, we adopted the CRG formulation and implemented graph construction using the Chemprop 2.0 ~\cite{chemprop} featurizer module. Atom mappings required for CRG generation were produced for each dataset using NextMove HazELNut ~\cite{hazelnut}. This preprocessing step enabled direct incorporation of reactant-product correspondence into a single graph structure. (Figure 1a)
\subsection{Molecule graph augmentation via subgraph removal} In graph contrastive learning, subgraph removal generates semantically consistent but structurally perturbed views; training encourages the model to learn invariances beyond local substructures. We use three masking-based views. For pretraining, an origin atom is selected uniformly at random, and masking expands outward to neighboring atoms using breadth-first-search, until 25\% of the atoms in the full CRG are masked.
For perturbation studies, one view restricts masking to reaction-center atoms, continuing until either 25\% of all atoms or all reaction-center atoms have been masked in the full CRG. The other view restricts masking to atoms outside the reaction center, i.e., side-chain atoms, until 25\% of the atoms in the full CRG are masked. After masking, all bonds adjacent to masked atoms are removed, resulting in a subgraph of the original molecular graph.

\begin{figure*}[t] \centering \includegraphics[width=\textwidth]{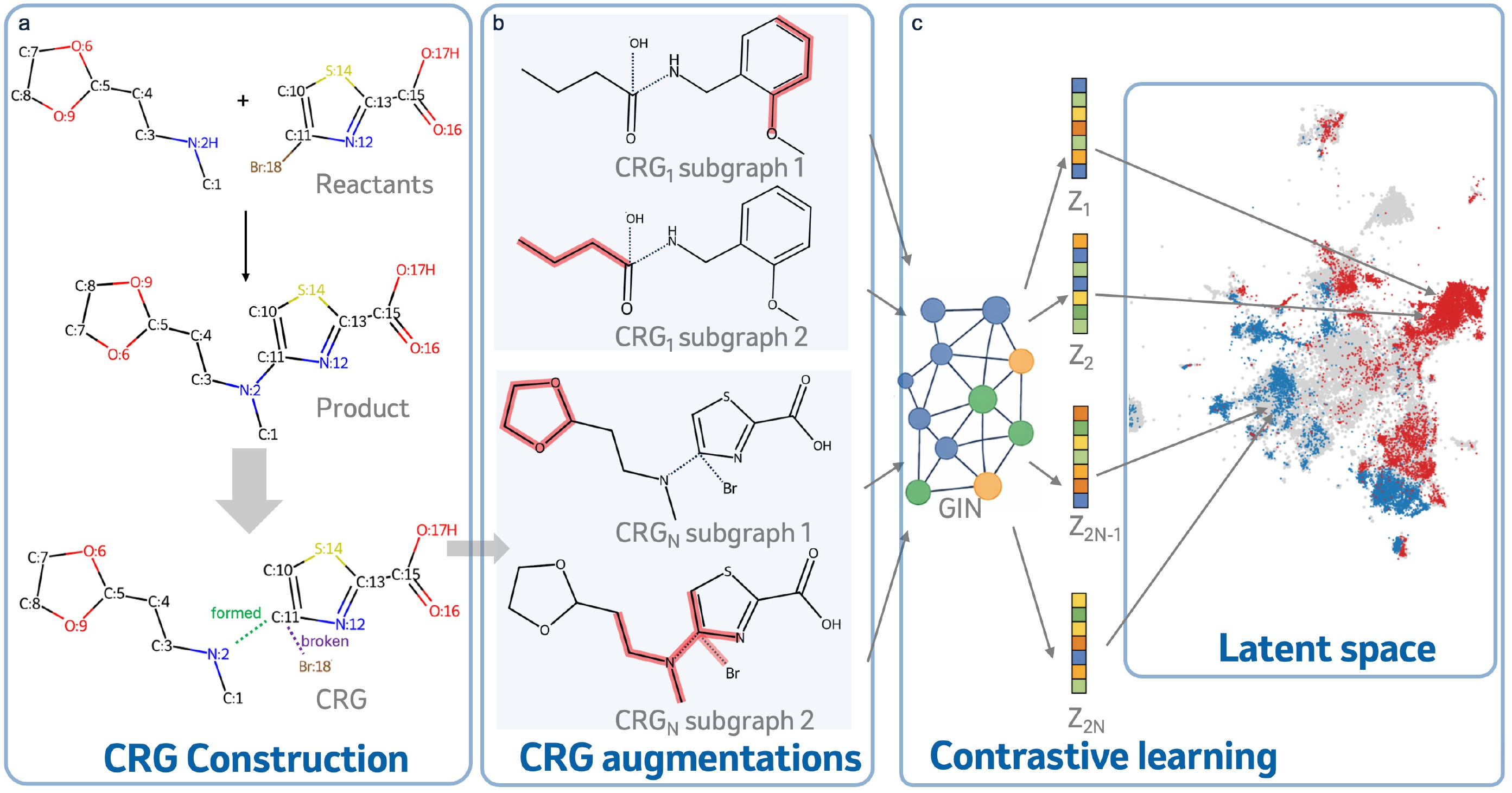} \caption{ Overview of RxnCLF training. The process consists of three stages. a: A reaction is converted from atom-mapped reactants and products into a CRG, which explicitly encodes bond formation/cleavage, and conserved substructures. Thin arrow represents reaction, while broad grey arrow represents CRG transformation. b: Two augmented CRG views are generated through subgraph augmentation, with masked atoms and deleted bonds shown in red. c: Contrastive pretraining with a shared GIN encoder enables the model to learn a compact reaction latent space of RxnCLF. Amide formation reactions are colored with red and C-N coupling reactions are colored with blue.  } \Description{verview of RxnCLF training. The process consists of three stages. a: A reaction is converted from atom-mapped reactants and products into a CRG, which explicitly encodes bond formation/cleavage, and conserved substructures. Thin arrow represents reaction, while broad grey arrow represents CRG transformation. b: Two augmented CRG views are generated through subgraph augmentation, with masked atoms and deleted bonds shown in red. c: Contrastive pretraining with a shared GIN encoder enables the model to learn a compact reaction latent space of RxnCLF. Amide formation reactions are colored with red and C-N coupling reactions are colored with blue.  } \label{fig:latent_space} \end{figure*}

\subsection{Contrastive learning for pretraining} Given a batch of $N$ atom-mapped reactions, we first construct a CRG for each reaction and generate two augmented graph views through subgraph augmentation. Views derived from the same reaction are treated as positive pairs, while views from different reactions serve as negative pairs. A Graph Isomorphism Network (GIN)~\cite{xu2018how} encoder maps each graph view to an embedding $h \in \mathbb{R}^{d}$, which is then projected by a fully connected layer into a latent representation $z$. We train the encoder using the Normalized Temperature-scaled Cross Entropy (NT-Xent) loss over the $2N$ projected views as follows~\cite{simclr}: \begin{equation} \ell_{i,j} = -\log \frac{ \exp \left( \operatorname{sim}(z_i,z_j)/\tau \right) }{ \sum_{k=1}^{2N} \mathbf{1}_{[k \neq i]} \exp \left( \operatorname{sim}(z_i,z_k)/\tau \right) } \label{eq:ntxent} \end{equation} where $\ell_{i,j}$ denotes the loss associated with a positive pair $(i,j)$, $\tau$ is a temperature hyperparameter, $\operatorname{sim}(\cdot,\cdot)$ denotes cosine similarity and $\mathbf{1}(\cdot)$ is the indicator function. This objective encourages agreement between positive pairs while pushing apart representations of negative pairs in the latent space.
\subsection{Reaction latent space evaluation}We compare three reaction representations: FP Difference (Morgan fingerprint, radius 2, 2048 bits)~\cite{drfp}, RxnFP (256 bits)~\cite{rxnfp}, and RxnCLF embedding (512 bits). FP Difference is computed by subtracting the hashed Morgan fingerprint of the products from that of the reactants. RxnFP is the latent embedding from BERT pretraining, and computed using the official implementation released by the authors. RxnCLF embedding is produced by our pretrained model. To assess the quality of the latent space, we use cosine similarity-based retrieval and report three metrics: K-Nearest Neighbor (KNN) stability, hubness, and Mean Reciprocal Rank (MRR). These metrics quantify local neighborhood consistency, embedding concentration, and ranking quality, respectively.
\begin{itemize} \item \textbf{KNN stability:} We randomly sampled $N$ ($N=50{,}000$) query reactions from Pistachio and computed their FP Difference, RxnFP, and RxnCLF embeddings. For each query, we retrieved the top 20 nearest neighbors using the original embedding and a perturbed version with added Gaussian noise. Stability was defined as the overlap ratio between the two neighbor sets $(S)$: \begin{equation} \frac{\left|S_{\mathrm{original}} \cap S_{\mathrm{perturbed}}\right|}{N} \end{equation} 
Higher KNN stability indicates a smoother and more stable latent space, where local neighborhood relationships are consistently preserved despite small perturbations.
\item \textbf{Hubness:} We measured hubness by counting how often each reaction appeared in the top-20 nearest neighbors of all queries under the three embeddings using the sampled data from the KNN stability experiment. For a sampled dataset \begin{equation} D=\{x_1,\ldots,x_N\}, \end{equation} let $NN_k(x_j)$ denote the $k$-nearest-neighbor set of $x_j$. The $k$-occurrence of sample $x_i$ is defined as~\cite{hubness},
\begin{equation} N_k(x_i) = \sum_{j=1}^{N} \mathbf{1} \left( x_i \in NN_k(x_j) \right), \end{equation} where $\mathbf{1}(\cdot)$ is the indicator function. Hubness is quantified by the skewness of the $k$-occurrence distribution: \begin{equation} \mathrm{Hubness} = S_{N_k} = \frac{ \mathbb{E} \left[ \left( N_k-\mu_{N_k} \right)^3 \right] }{ \sigma_{N_k}^{3} }. \end{equation} where $\mu_{N_k}$ and $\sigma_{N_k}$ denote the mean and standard deviation of the $k$-occurrence distribution, respectively, and $k=20$. Higher hubness indicates more pronounced hub formation and a less uniform neighborhood structure.
\end{itemize}
\begin{itemize} \item \textbf{Mean Reciprocal Rank (MRR):} To quantify retrieval quality, we computed the mean reciprocal rank (MRR) across 50,000 query reactions.

\begin{equation}
\mathrm{MRR}
=
\frac{1}{N}
\sum_{i=1}^{N}
\frac{1}{\mathrm{rank}_i},
\end{equation}

where $N$ is the number of query reactions and
$\mathrm{rank}_i$ denotes the rank position of the first retrieved reaction sharing the same reaction as the $i$-th query. Higher MRR indicates that relevant reactions are retrieved closer to the top of the ranked list, reflecting a more informative and better-aligned latent space.
\end{itemize}

\begin{itemize} \item \textbf{Query Retrieval-based Analysis:}
We randomly sampled 500 query reactions from Pistachio. For each query, we retrieved the top-50 nearest neighbors based on embeddings from the three latent spaces. We then evaluated retrieval quality using a top-$k$ protocol ($k \in [1,50]$) by computing the fraction of retrieved reactions sharing the same reaction superclass/type as the query:

\begin{equation}
\frac{1}{k}
\sum_{r \in R_k}
1\!\left(
\mathrm{type}(r), t
\right),
\end{equation}

where $R_k$ denotes the set of top-$k$ retrieved reactions and $t$ denotes the target reaction superclass/type, $ type(r)$ returns the superclass/type label of r, and $\mathbf{1}(\cdot)$ denotes the indicator function. This ratio assesses how well the embeddings preserve reaction-type information. Additionally, Tanimoto similarity between products of query and retrieved reaction was computed using count Morgan fingerprints for both full molecule and Bemis-Murcko scaffold to aid interpretation of the retrieval results ~\cite{riniker2013}.
\end{itemize}

\subsection{Out-of-distribution analysis} Beyond Pistachio training dataset, Pd-catalyzed BH coupling and proprietary amide formation datasets were projected into the learned latent space. A query of retrieval analysis was then conducted by retrieving similar Pistachio reactions for each dataset.

\subsection{Perturbation analysis} We compared Pistachio embeddings produced for each reaction, from three graph variants: (1) the full reaction graph, (2) a reaction subgraph with the reaction center removed, and (3) a subgraph produced by randomly removing side chains. Pairwise cosine similarity, Pearson similarity, Euclidean distance and Manhattan distance were computed among the three corresponding embeddings to quantify how these structural perturbations affect the latent representations.
\subsection{Yield prediction} We evaluate four models Chemprop 2.0, GIN, YieldBERT, and RxnCLF across four yield prediction datasets. Chemprop 2.0 and GIN were trained from scratch using only the downstream yield prediction data, while YieldBERT and RxnCLF were initialized with pretrained representations learned from the USPTO and Pistachio reaction corpus, respectively. Chemprop 2.0 is used with its default configuration on CRG inputs. The GIN is trained directly on CRG representations. YieldBERT is implemented using its pretrained weights adopted from publication and finetuned with prediction datasets. To leverage the pretrained latent representations learned by RxnCLF, we finetuned the pretrained encoder with a prediction head. The encoder is initialized from the RxnCLF weights, while the prediction head is randomly initialized. For the dataset containing diverse reagents, their information is encoded as categorical features and embedded into the prediction head. The entire model is then optimized end-to-end in a supervised manner on the downstream yield prediction task using the standard Mean Squared Error (MSE). For each dataset, we split reactions into 80\%/5\%/15\% train/validation/test sets, repeating five times with distinct random seeds. Early stopping is controlled by validation loss. Model performance is recorded by $R^2$ on the test set.

\subsection{Datasets}For pretraining, we use 1.7M Pistachio reactions after removing duplicate, unassigned reactions, and those with multiple products ~\cite{pistachio}. For downstream yield predictions, we evaluate four datasets spanning both curated academic and industrial benchmarks. The first is the Buchwald--Hartwig dataset from prior work ~\cite{bh_dataset}, which contains 4,608 reactions, generated from 15 aryl and heteroaryl halides, 4 Buchwald ligands, 3 bases, and 23 isoxazole additives. The second is the Pd-catalyzed BH coupling dataset~\cite{cn_dataset}, which contains 4,088 reactions constructed from 342 halides and 348 amines under fixed catalyst and base settings. The third is a proprietary C--N coupling dataset with around 29k reactions covering large reaction space and sparse reagent combinations (including Buchwald--Hartwig, copper-catalyzed Goldberg reaction, Ullmann reaction, SNAr, and etc); due to the large number of missing reagent annotations, reagents were excluded from fine-tuning, and the label is conversion. The fourth is an proprietary amide formation dataset with around 26k reactions covering diverse reactants, 7 bases, 5 solvents, and 12 additional reagents. Together, the two proprietary datasets provide a more realistic industrial setting for evaluating model generalization. 

\section{Results}
\subsection{Reaction latent space analysis}
\subsubsection{Reaction latent space visualization:}We first evaluate whether RxnCLF learns a reaction latent space that is more structured and search-friendly than existing representations. To this end, we project Pistachio reactions into three reaction spaces: (1) FP Difference, (2) RxnFP, and (3) RxnCLF embeddings. For visualization, we first reduce the dimension to 50 using PCA and then to 2 using UMAP (Figure 2).

The projection results reveal clear differences across the three representations. FP Difference is widely dispersed, suggesting limited local structure and substantial heterogeneity in the latent space. RxnFP forms a scattered and fragmented space with only sparse superclass-level clustering, indicating weaker preservation of local neighborhood structure and, consequently, reduced interpretability. Nevertheless, the fragmented clustering of reactions from the same superclass still suggests that RxnFP captures coarse reaction-class semantics. In contrast, RxnCLF embeddings provide the tightest and most structured latent space: reactions from the same superclass tend to occupy common regions, indicating that the latent space is sensitive to reaction category change. 

\begin{figure*}[t] \centering \includegraphics[width=\textwidth]{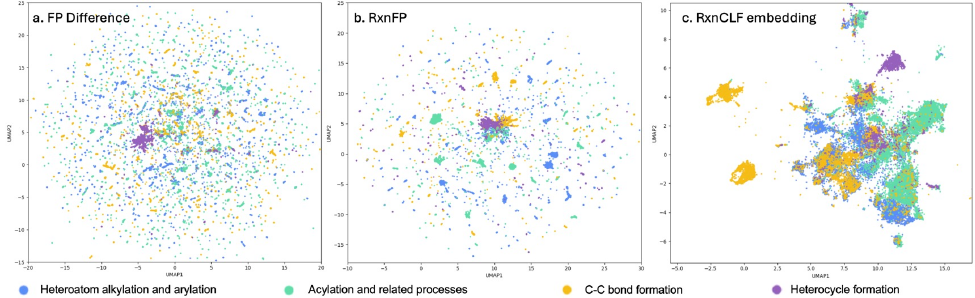} \caption{ Visualization of the three latent spaces, colored by the four most frequent HazELNut reaction superclasses. a. FP Difference; b. RxnFP; c. RxnCLF embeddings. } \Description{ Visualization of the three latent spaces, colored by the four most frequent HazELNut reaction superclasses. a. FP Difference; b. RxnFP; c. RxnCLF embeddings. } \label{fig:latent_space} \end{figure*}

\subsubsection{Latent space retrieval analysis:} A high-quality latent space should support reliable similarity search in addition to being structured. We therefore evaluated KNN stability, hubness, MRR, and embedding extraction cost for the three representations (Table 1). RxnFP and RxnCLF embeddings both achieve KNN stability above 0.99, indicating strong local consistency in the latent space, where semantically similar reactions are consistently mapped to nearby regions. RxnCLF exhibits slightly higher hubness, but also achieves the highest MRR (0.986), reflecting superior retrieval quality. Along with the fast extraction speed, RxnCLF is the most effective among the tested models for similarity search and potentially for downstream use.

\begin{table}
\caption{Comparison of FP Difference, RxnFP, and RxnCLF embeddings in latent space retrieval quality and extraction cost.} \label{tab:embedding_comparison} 
\centering \begin{tabular}{lccc} \toprule Metric & FP Difference & RxnFP & RxnCLF \\ \midrule KNN Stability & 0.912 & 0.999 & 0.998 \\ Hubness & 3.34 & 1.26 & 1.88 \\ MRR & 0.673 & 0.959 & 0.986 \\ Extraction Cost & 30 min / 2M & 390 min / 2M & 15 min / 2M \\ \bottomrule \end{tabular} \Description{ Comparison of FP Difference, RxnFP, and RxnCLF embeddings across four retrieval-related metrics. KNN Stability and MRR are higher-is-better metrics, while Hubness measures the skewness of neighbor occurrence distributions. Extraction Cost reports the approximate computational time required to generate embeddings for two million reactions. } \end{table}

\subsubsection{Query-based retrieval analysis:} To probe retrieval behavior more directly, we conducted 500 query-based searches and measured the Tanimoto similarity between each query product and its retrieved neighbors (Figure 3). RxnCLF consistently returns neighbors with the highest product and scaffold similarities, indicating that it preserves both reaction-center features and broader structural context. By contrast, RxnFP achieves the lowest product similarity overall. The slightly higher Bemis–Murcko-based similarity for RxnCLF further suggests that its latent space better preserves scaffold-level structure while remaining sensitive to fine-grained side-chain variation.

Second, we evaluated the fraction of retrieved reactions that share the same reaction superclass and reaction type as the query. RxnFP achieves the best performance on this task: the same-type fraction exceeds 90\% within the top 50 retrieved results, while both RxnCLF and FP Difference show markedly lower same-type fractions. Since reaction type is a sub-label within the superclass, its retrieval fraction is expected to be lower than the corresponding superclass-level fraction. These results suggest that RxnFP places stronger emphasis on reaction-center features that define reaction type, whereas RxnCLF embeddings preserve more structural-context information at the expense of class purity.

\begin{figure}[t] \centering \includegraphics[width=\linewidth]{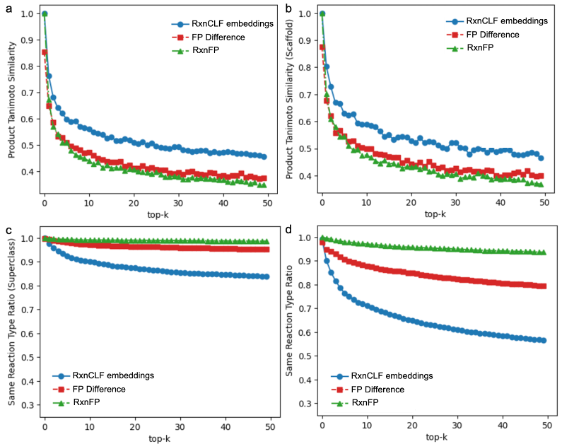} \caption{ Evaluation of reactions retrieved from the three latent spaces relative to the query reaction, using, a. product Tanimoto similarity based on Morgan count fingerprints, b. product Tanimoto similarity based on Bemis-Murcko scaffold plus Morgan count fingerprints and c. same reaction type ratio, d. same reaction superclass ratio.  } \Description{ Three latent-space visualizations comparing FP Difference, RxnFP, and RxnCLF embeddings. Colors indicate reaction superclasses and illustrate differences in cluster structure and class separation among the three representations. } \label{fig:latent_space} \end{figure}

An illustrative example (Figure 4) shows one query and the top 5 retrieved reactions from each space: comparing the search results, RxnCLF can retrieve the most similar reactions with highest product similarity at both molecule and scaffold level. RxnCLF retrieved top 5 reactions include a reaction labeled Ester Schotten-Baumann (top-4) while the query is a Williamson ether synthesis. Although the reaction types differ, the retrieved product is visually and chemically similar to the query product, both featuring a benzene ring and a heterocycle ring. This result suggests that RxnCLF can serve as a useful search engine for chemists to retrieve structurally and mechanistically related reactions beyond exact reaction types.

\begin{figure*}[t] \centering \includegraphics[width=\textwidth]{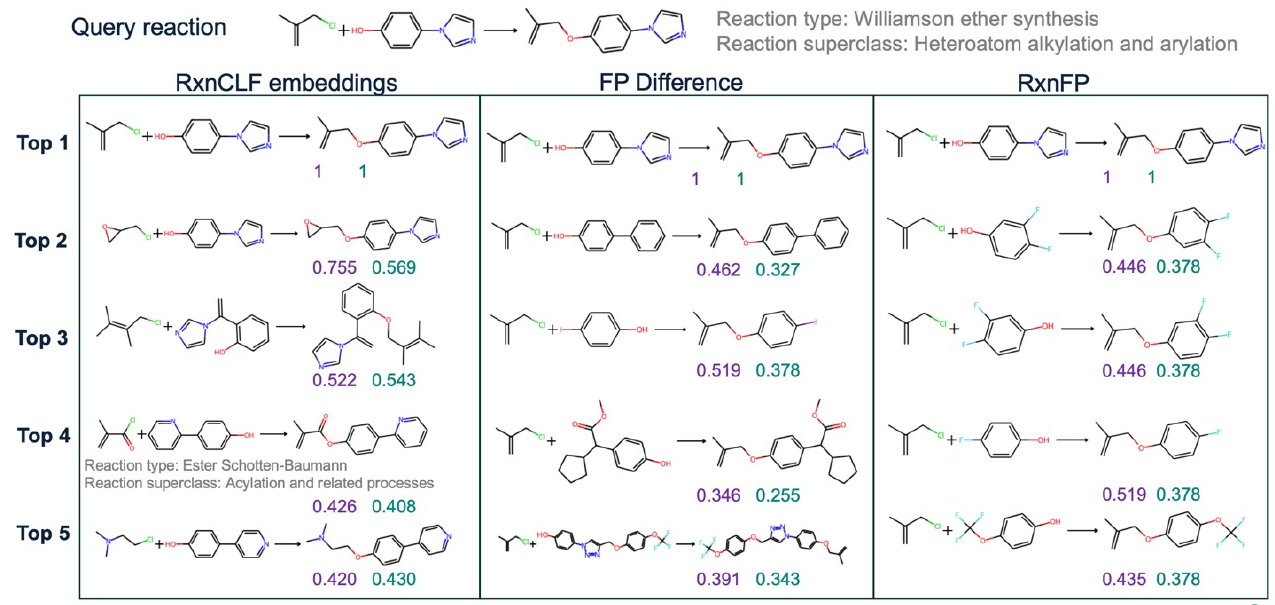} \caption{ Top-5 similar reactions retrieved from three latent spaces for a given query reaction. Product similarity scores are shown using two color schemes: purple indicates the Tanimoto similarity based on the Morgan count fingerprint, and green indicates the Tanimoto similarity based on the Bemis–Murcko scaffold plus Morgan count fingerprint. } \Description{ Three latent-space visualizations comparing FP Difference, RxnFP, and RxnCLF embeddings. Colors indicate reaction superclasses and illustrate differences in cluster structure and class separation among the three representations. } \label{fig:latent_space} \end{figure*}

Taken together, these results highlight complementary strengths across the three representations. RxnFP clusters reactions strongly by reaction centers and reaction types, making it effective for coarse-grained categorization. The FP Difference captures local reaction-center changes but remains relatively scattered, limiting its utility for reliable similar reaction search. In contrast, RxnCLF learns a compact latent space that preserves both reaction-center structure and product-level similarity, making it particularly well suited for retrieval. This behavior is also directly relevant to downstream tasks, especially yield prediction. Unlike reaction classification, yield depends on both the reaction type and substrate context. By retaining reaction-center information while preserving additional structural variation, RxnCLF provides a balanced representation for this task, which helps explain its superior performance on downstream prediction tasks (section 3.2).

\subsubsection{OOD analysis and latent space generalization:} We projected out-of-distribution (yield prediction) datasets into the RxnCLF embeddings trained from Pistachio. These sets included the Pd-catalyzed BH coupling dataset, proprietary HTE C-N coupling and amide formation dataset, containing reactions not present in Pistachio, and a Buchwald-Hartwig set of which 3 out of 15 reaction substrates are included in Pistachio. In the UMAP projections (Figure 5) each external set forms local clusters that overlap the regions established from Pistachio pretraining, indicating that the RxnCLF space places out of distribution examples into the same class specific neighborhoods rather than isolating them as artifacts. Overall, the visualization suggests that the RxnCLF latent space organizes unseen reactions into chemically meaningful neighborhoods.

\begin{figure}[t] \centering \includegraphics[width=\linewidth]{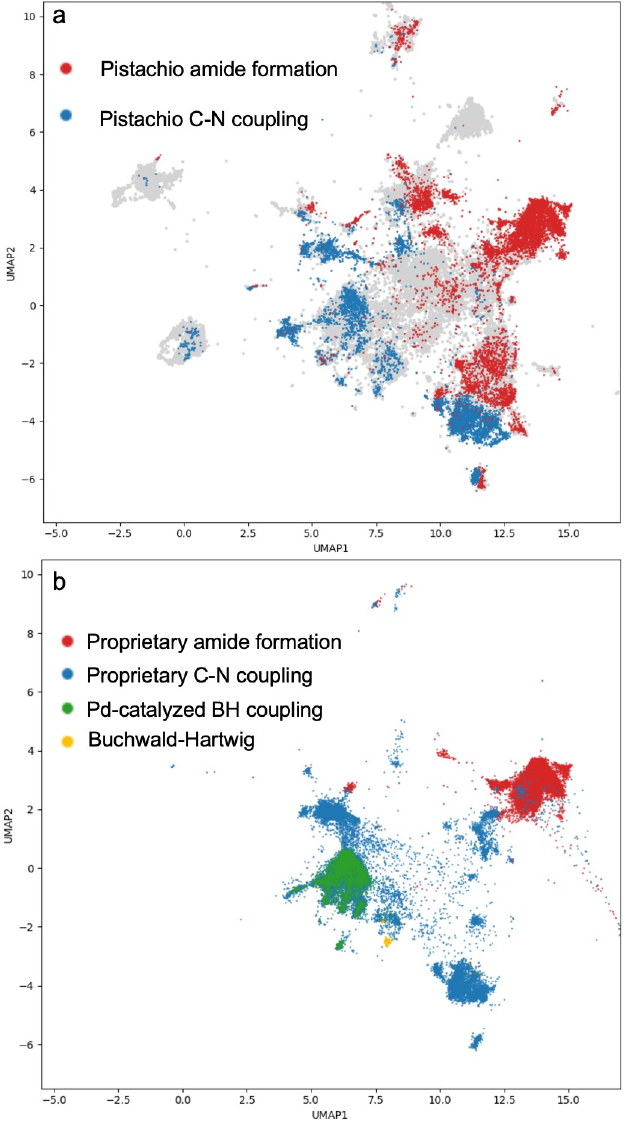} \caption{ Latent space visualization of, a. the in-distribution Pistachio reactions and b. the out-of-distribution datasets. } \Description{ Three latent-space visualizations comparing FP Difference, RxnFP, and RxnCLF embeddings. Colors indicate reaction superclasses and illustrate differences in cluster structure and class separation among the three representations. } \label{fig:latent_space} \end{figure}

Next, we evaluated retrieval behavior by using the Pd-catalyzed BH coupling and proprietary amide formation datasets as queries against the Pistachio search database. Retrieval quality, measured by product Tanimoto similarity and the fraction of retrieved reactions sharing the same reaction type as the query, decays smoothly with increasing rank (Figure 6). The monotonic decay indicates that the latent space is continuous and well structured, such that nearby points correspond to similar products and, when relevant, to the same reaction types. Among the two external datasets, the amide formation queries show higher product similarity and same-type ratios than the C–N queries, suggesting that the amide formation reactions are more similar to the Pistachio training distribution. These results demonstrate that RxnCLF preserves its performance under distribution shift.

\begin{figure}[t] \centering \includegraphics[width=\linewidth]{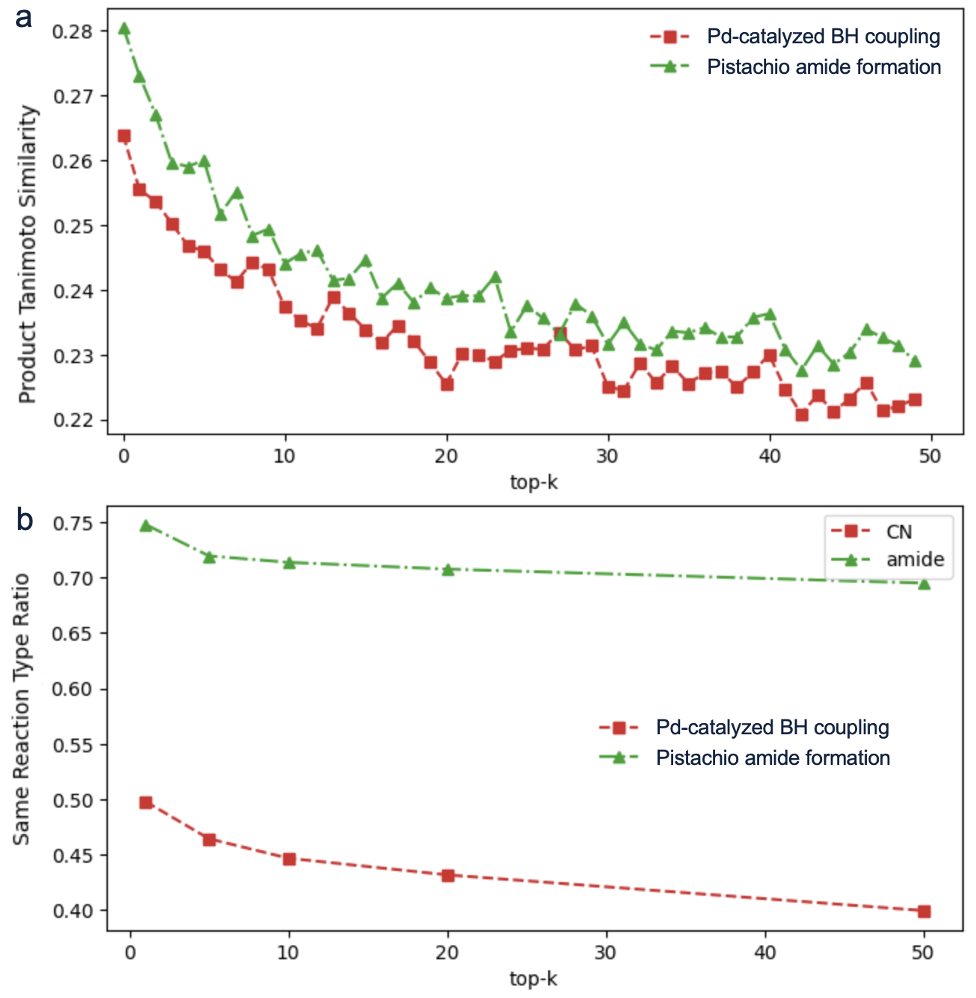} \caption{ OOD evaluation: a. product Tanimoto similarity between external dataset queries and retrieved reactions in the Pistachio latent space; b. same-reaction-type ratio between external dataset queries and retrieved reactions in the Pistachio latent space.} \Description{ Three latent-space visualizations comparing FP Difference, RxnFP, and RxnCLF embeddings. Colors indicate reaction superclasses and illustrate differences in cluster structure and class separation among the three representations. } \label{fig:latent_space} \end{figure}

\subsubsection{Latent space perturbation analysis:} we assess latent-space sensitivity by applying controlled perturbations to reaction graphs and comparing the resulting embeddings. For each reaction, we construct three variants: the full graph (A), a subgraph with the reaction center removed (B), and a subgraph with randomly removed side chain atoms (C). As shown in Table 2, all pairwise similarities remain consistently high, with cosine similarity and Pearson correlation exceeding 0.99, indicating that CRG learns a smooth and stable latent space under moderate structural perturbations.

\begin{table}[t] \caption{Pairwise comparison of the three latent spaces, evaluated using similarity and distance metrics for the full graph (A), the reaction-center-removed subgraph (B), and the side-chain-removed subgraph (C).} \label{tab:view_similarity} \centering \resizebox{\linewidth}{!}{ \begin{tabular}{lccc} \toprule Metric & $\mathrm{Sim}(A,B)$ & $\mathrm{Sim}(A,C)$ & $\mathrm{Sim}(B,C)$ \\ \midrule Cosine Similarity & $0.993 \pm 0.005$ & $0.995 \pm 0.004$ & $0.991 \pm 0.006$ \\ Pearson Correlation & $0.993 \pm 0.005$ & $0.995 \pm 0.005$ & $0.991 \pm 0.006$ \\ Euclidean Distance & $4.75 \pm 1.48$ & $3.93 \pm 1.43$ & $5.21 \pm 1.83$ \\ Manhattan Distance & $76.47 \pm 24.00$ & $64.56 \pm 23.41$ & $85.38 \pm 30.38$ \\ \bottomrule \end{tabular} } \Description{ Pairwise similarity and distance statistics between latent representations obtained from three augmented views. } \end{table}

Notably, the full graph (A) and the side-chain-removed subgraph (C) exhibit the highest similarity, whereas the reaction-center-removed subgraph (B) shows the largest deviation from the others. This gap suggests that the reaction center contributes more substantially to the learned representation than side-chain atoms, confirming that CRG encodes reaction transformations in a chemically meaningful and structurally aware manner. These results highlight the mechanistic advantage of CRG: by distinguishing reaction centers from side chain, it captures both robustness to minor perturbations and sensitivity to transformation-critical substructures.

\begin{table*}[!t] \caption{ ($R^2$) comparison across multiple reaction datasets and models for yield prediction (Results are reported as mean $\pm$ standard deviation).} \label{tab:yield_prediction} \centering \begin{tabular}{lcccc} \toprule Dataset & Chemprop & GIN & RxnCLF & YieldBERT \\ \midrule Buchwald--Hartwig (BH) & $0.617 \pm 0.083$ & $0.930 \pm 0.012$ & $\mathbf{0.962 \pm 0.003}$ & $0.951 \pm 0.005$ \\ Pd-catalyzed BH coupling & $0.726 \pm 0.010$ & $0.597 \pm 0.027$ & $\mathbf{0.769 \pm 0.016}$ & $0.659 \pm 0.018$ \\ Proprietary C--N coupling & $0.469 \pm 0.032$ & $0.471 \pm 0.017$ & $\mathbf{0.484 \pm 0.020}$ & $0.397 \pm 0.016$ \\ Proprietary amide formation & $0.395 \pm 0.052$ & $0.358 \pm 0.016$ & $\mathbf{0.428 \pm 0.024}$ & $0.418 \pm 0.012$ \\ \bottomrule \end{tabular} \Description{ Comparison of reaction yield prediction performance across Chemprop, GIN, RxnCLF, and YieldBERT on four public and proprietary datasets. Values are reported as mean and standard deviation of $R^2$. RxnCLF achieves the highest score on all four datasets. } \end{table*}
 
Collectively, the OOD projections, search behavior, and perturbation experiments demonstrate that the RxnCLF latent space is continuous, stable under small input changes, and organized such that external reactions are embedded into meaningful, class consistent neighborhoods.

\subsection{Yield prediction}

The RxnCLF was fine-tuned for yield prediction on four datasets, respectively: Buchwald-Hartwig, Pd-catalyzed BH coupling, proprietary C-N coupling, and amide formation. Buchwald-Hartwig dataset is a standard benchmark with few substrate variations for multiple conditions. Pd-catalyzed BH coupling dataset represents industrial library optimization with a set of substrates. Proprietary C-N coupling and amide formation datasets mined from HTE data reflect more sparse experimental conditions. RxnCLF achieved the highest $R^2$ across all datasets (Table 3). In addition, pretrained RxnCLF consistently improved performance relative to GIN models trained from scratch, indicating efficient use of representation learned from foundation models.

As shown in Figure 5b, public datasets (Buchwald-Hartwig and Pd-catalyzed BH coupling) occupy a narrow but well-controlled region of reaction space, typically varying a small number of precursors around one or a few reaction templates. Despite their limited scope, they provide high-quality supervision because reactions are recorded uniformly and yields are measured with the same analytical procedures. In contrast, industrial data cover a broader chemical space but is noisier. For instance, proprietary C-N coupling reaction dataset includes a variety of transition-metal catalyzed reactions including Pd-catalyzed BH coupling, copper-catalyzed Ullmann and Goldberg coupling, SNAr and other types. Heterogeneity of reaction types in the data might lead to poorer performance. In addition, internal experiments are conducted by different scientists; recorded reactions could miss key reagents or conditions due to human errors. This tradeoff in data quality and coverage likely explains why RxnCLF works better on public curated over industry data.

\section{Conclusions}
We presented RxnCLF, a contrastive learning framework built on the condensed reaction graph for reaction representation learning. By encoding reaction transformation structure, RxnCLF learns a chemically interpretable and transformation-aware latent space that captures both reaction-center information and broader side chain context. Pretraining on large-scale reaction data enables the model to learn enriched representations and robust reaction embeddings under sparse supervision.

Our results show that RxnCLF learns a structured and search-capable latent space. This translates into improved downstream yield prediction across tested benchmarks. Along with the embeddings which preserves meaningful similarity, these findings demonstrate the value of explicit transformation modeling for scalable and transferable reaction representation learning.

\bibliographystyle{unsrt}
\bibliography{references}

\end{document}